\documentclass[10pt,twocolumn,letterpaper]{article}
\usepackage{iccv}
\usepackage{times}
\usepackage{epsfig}
\usepackage{graphicx}
\usepackage{amsmath}
\usepackage{amssymb}
\usepackage{dsfont}
\usepackage{tablefootnote}
\usepackage{multirow}
\usepackage[ruled]{algorithm2e}

\usepackage[breaklinks=true,bookmarks=false]{hyperref}

\iccvfinalcopy 
\def\iccvPaperID{****} 

\newcommand{\norm}[1]{\left \lVert #1 \right \rVert_{F}^2}

\DeclareMathOperator*{\argmin}{arg\,min}
\DeclareMathOperator*{\minimize}{min}
\begin{document}

\title{Semi-supervised Zero-Shot Learning by a Clustering-based Approach}
 \author{Seyed Mohsen Shojaee and Mahdieh Soleymani Baghshah\\
 Sharif University of Technology \\
 Tehran, Iran\\
 {\tt\small mshojaee@ce.sharif.edu, soleymani@sharif.edu}
 }

\maketitle

\begin{abstract}
In some of object recognition problems, labeled data may not be available for all categories.
 Zero-shot learning utilizes auxiliary information (also called signatures)
 describing each category in order to find a classifier that can recognize samples
from categories with no labeled instance. 
In this paper, we propose a novel semi-supervised zero-shot learning method that works on an embedding space corresponding to
abstract deep visual features. We seek a linear transformation on signatures to map them onto the visual features,
such that the mapped signatures of the seen classes are close to labeled samples of the corresponding
classes and unlabeled data are also close to the mapped signatures of one of the unseen classes.
 We use the idea that the rich deep visual features provide a representation
 space in which samples of each class are usually condensed in a cluster. The effectiveness of the proposed method is demonstrated through extensive
experiments on four public benchmarks improving the state-of-the-art prediction accuracy on three of them.
\end{abstract}

\section{Introduction}
Zero-shot learning \cite{bengio08, hinton09,lampert09,farhadi09} is an extension to the conventional supervised learning scenario
that does not need labeled instances for all categories in order to recognize them.
Instead, some sort of description that is called \textit{class signatures} is also available for all the categories.
Signatures may be a set of human-annotated discriminative attributes or textual description of the categories.
The problem addressed by zero-shot learning rises naturally in practice wherever it is not feasible to acquire abundant labeled instances for all the categories (e.g., fine-grained classification problems).
To describe the task more precisely, in the training phase, labeled instances for some categories which are called seen classes are provided
while for other categories called unseen ones there is no labeled instance available.
In the test phase, unlabeled instances should be classified into seen or unseen categories. In this work, however, we focus on the most popular version of zero-shot recognition in which test instances belong only to unseen categories.

Most existing methods for zero-shot learning focus on using labeled images to learn a compatibility function indicating how similar an image is
to each label embedding~\cite{Akata2015,emb15,sse}. Each instance will then be labeled with the category having the most compatible signature.
 On the other hand, recent advances in deep convoloutional neural networks provide rich visual features with high discrimination capability~\cite{vgg}.
  We will show in Section\ref{experiments} through experiments that the space of deep visual features is indeed a rich space in which instances of different categories usually form natural clusters. However, little attention has been paid to exploiting this property of visual features in the context of zero-shot learning.


In this paper, we propose a semi-supervised zero-shot learning method that uses both labeled samples of seen classes and unlabeled instances of unseen classes to find a more proper representation of labels (i.e., label embedding) in the space of deep visual features. We seek a linear transformation to map the auxiliary information to the space of abstract visual features and jointly find assignments of unlabeled samples to unseen classes.
We intend to learn a linear transformation such that the mapped signature of each seen class tends to be representative for samples of the corresponding class and simultaneously it is possible to find an assignment of unlabeled samples to unseen classes such that the mapped signature of each unseen class will also tend to be the representative of the assigned samples to that class.
Using the unlabeled samples of the unseen classes, we can substantially mitigate the domain shift problem previously introduced in \cite{eccv14} that impairs the zero-shot recognition performance.
We also propose a simpler method that does not jointly learn the linear transformation on class signatures and label assignments to unlabeled data.
Instead, after finding the mapping according to just instances of seen classes, it uses a clustering algorithm to assign labels to instances of unseen classes. 

In Section~\ref{experiments}, we present experimental results on four popular zero-shot classification benchmarks and see that the proposed method outperforms the state-of-the-art methods on three out of these four datasets.

The rest of paper is organized as follow. First, in Section~\ref{related}, we briefly introduce existing methods for zero-shot learning.
Then, in Section~\ref{proposed}, we present our semi-supervised zero-shot learning method. In Section \ref{experiments}, we report our experiments and finally in Section~\ref{conclusion}
we conclude.

\section{Related Work} \label{related}
Most existing methods for zero-shot object recognition can be described as finding a compatibility function scoring how
similar an image and a description are.
We can consider the following steps for these methods:
\begin{itemize}
  \item Find (or use the existing) embeddings for class labels in a semantic space.
  \item Map images into that semantic space.
  \item Classify images in the semantic space based on the compatibility of the mapped image and the embedded labels in this space (usually using a nearest neighbor classifier or label propagation).
\end{itemize}
Learning of these three steps may be done independently or jointly.

A notable body of work in zero-shot recognition belongs to attribute prediction from images \cite{lampert09, topicmodel, ajoint11, unified13, suzuki14}.
In these methods, the semantic label embeddings are considered to be externally provided attributes as auxiliary information. Thus, attributes as label embeddings are available and the task is just to map images to the semantic space, i.e., predicting attributes for the images.
Early methods, like \cite{lampert09}, assume independence between attributes and train binary attribute classifiers.
Probabilistic graphical models have been utilized to model and/or learn correlations among different attributes \cite{topicmodel, unified13} to improve the prediction of the attributes.
In \cite{jayaraman14}, a random forest approach has been employed that accounts for unreliability in attribute predictions for final class assignment.
In \cite{Akata2015pami}, a max-margin objective function similar to the structured SVM is defined for attribute-based image classification.

More recent works exploit bilinear models \cite{Yu2013, devise, convex, sse, emb15,semi15} that are also equivalent to embedding images and labels into a
common space and considering the inner product in the embedded space as the compatibility score.
Until now, several objective functions have been proposed for learning such bilinear models.
In \cite{emb15}, the sum of the squared error on the label prediction is used.
However, extra regularization terms that compensate undesirable characteristics of this cost function are also utilized.
This method can be seen as learning a mapping that transforms description of each class to a linear classifier for that class.
This idea has also been used in \cite{li15max, semi15} that introduce a max margin objective function for this purpose.
 These two methods also learn labels for test instances simultaneously and so they differ from almost all of other existing methods in this way. This provides the possibility of
leveraging unsupervised information available in test images, for instance as done in \cite{semi15}, by using a Laplacian regularization term that penalizes similar objects assigned to different classes.

Designing label embeddings in multi-class classification
is another line of research that can also be used for zero-shot recognition.
 In \cite{Yu2013}, an objective function is proposed to derive such label embeddings based on information about similarities among categories.
A relatively popular embedding for labels is to describe unseen categories as how similar they are to the seen ones.
One way to use this embedding is creating classifiers for unseen categories by linear combination of
classifiers for seen categories using similarity scores as mixing weights.
In \cite{convex}, the outputs from the softmax layer of a CNN trained on seen categories are used to score similarity between test instances and seen classes.
Using these outputs as weights, the introduced method in \cite{convex} represents images in the semantic space as a convex combination of seen class label embeddings.
Moreover, in \cite{sse}, a histogram showing seen class proportions is used for label embedding and then a max margin framework is defined to embed images in this space. The authors of \cite{convex} extend their work further in \cite{agnostic} and formulate a supervised dictionary learning method that jointly learns image and label embeddings.
 The idea of combining already available classifiers to create new ones for unseen categories is also used in \cite{Synthesized}
 but rather than using seen categories as basis, they define a set of (possibly smaller) \textit{phantom} classes and learn base classifiers on them.

 Although most of the studies on zero-shot recognition consider attributes as auxiliary information, some of the existing methods utilize textual
  information for classes as auxiliary information.
  This text be obtained from online encyclopedias or be just the name of classes.
   Some existing methods first extract features from auxiliary text information and then turn them into vectors that can be treated analogous to attribute vectors.
 \cite{devise} introduces a bilinear model to find the compatibility score of deep visual features and Word2vec \cite{word2vec} representation of class names. \cite{ba2015} proposes nonlinear mappings modeled by neural networks on the image and the text inputs to find their compatibility.
  \cite{mohamed13} presents an objective function to predict classifier parameters from textual descriptions. In \cite{Akata2015}, different label embeddings such as attribute vectors, GloVe \cite{pennington2014glove}, word2vec \cite{word2vec}, a variant of word2vec with weak supervision, and also a combination of these different embeddings have been considered as the label embedding for zero-shot recognition. In \cite{Xian2016}, this work is extended further to model nonlinear compatibility
   functions that can be expressed as a mixture of bilinear models.
  In \cite{Qiao2016}, a modification of \cite{emb15} is presented as for use with textual auxiliary information by decomposing the bilinear mapping.

In \cite{Fu2016}, a set of vocabulary much larger than just seen and unseen class names is used and mapping from images to word embeddings is learned
by  maximizing the margin with respect to all words in the vocabulary; this framework can be used in zero-shot and also supervised and open set learning problems.
In  \cite{Akata2016}, authors propose to use multiple auxiliary information and also  part annotation in image domain to compensate for weaker supervision in textual data.
Convolutional and recurrent neural networks  have also been used for text embedding in \cite{Akata2016rnn}.

 The most related methods to our method are the introduced ones in \cite{li15max, semi15, Kodirov2015} that are indeed semi-supervised zero-shot learning methods. Here, we briefly specify the differences between these methods and ours. First, we use abstract visual features obtained by deep learning as the semantic space as opposed to these methods. \cite{li15max, semi15}
learn a max margin classifier on the image space classifying both seen and unseen instances while we use a ridge regression to map signatures to the semantic visual space resulting in a much simpler optimization problem to solve. Since samples of different classes are usually condensed in distinct regions of the deep visual representation space, our proposed optimization problem is based on clustering of data in this space and we try to map the class signatures on the centroid of the corresponding samples. We also explicitly account for domain shift problem in our objective function and thus achieving better results compared to these methods.

There are major differences between our work and ~\cite{Kodirov2015} using a dictionary learning scheme
in which coding coefficients are considered to be label embeddings in a semantic space
and a sparse coding objective is used to map images into this representation space.
Most importantly, in our method labels of unseen instances are jointly learned
with the mapping of the signatures to the semantic space in our objective function while in \cite{Kodirov2015}
the label prediction is accomplished using the nearest neighbor or the label propagation on embeddings of images.
Also, we do not need to learn embedding of test instances in the semantic space as opposed to \cite{Kodirov2015},
alternatively we learn just the representation of class signatures in the visual domain.

\section{Proposed Approach} \label{proposed}
In this section, we introduce a zero-shot learning method that uses the deep visual features as the semantic space and learns
a mapping from class signatures to this semantic space and also learns labels of instances belonging to unseen classes.
 First, we propose a simple and efficient semi-supervised zero-shot learning method in Section~\ref{clustering}.
  Then, we introduce an optimization problem that tries to simultaneously learn the mapping and the label assignment to test instances in Section~\ref{joint}.
   Finally, we introduce an iterative method to solve this optimization problem in Section~\ref{optimization}
   and use the simple method proposed in Section~\ref{clustering} to find a start point for this method (i.e., as an initial labellings for instances of unseen classes).
\subsection{Notation}
Let $X, \mathbf{x}$, and $x$ denote matrices, column vectors, and scalars respectively. $\norm{X}$ shows the squared Frobenius norm of a matrix and
$X_{(i)}$ denotes its $i$th column.
Suppose there are $n_s$ seen categories and $n_u$ unseen categories. For each category y,
auxiliary information $a_y \in \mathbb{R}^r$ is available. We assume that labels $\{1, \ldots, n_s \}$ correspond to seen categories.

Let $X_s \in \mathbb{R}^{d \times N_s}$ and $X_u \in \mathbb{R}^{d \times N_u}$
denote matrices whose columns are seen and unseen images respectively where $d$ is the dimension of image features.
$S_s = [a_1, \ldots, a_{n_s}]$ presents the matrix of signatures for seen classes. $S_u$ is also defined similarly for unseen classes.
$Z_s = [ \mathbf{z_1}, \ldots, \mathbf{z_{N_s}} ]$
contains labels of training data in one-hot encoding format.

\subsection{Clustering Method} \label{clustering}
Our first method can be roughly summarized in three steps:
\begin{enumerate}
  \item Using data from seen classes, we learn a linear mapping from attribute vectors to the semantic space.
  \item We find a data clustering using our proposed semi-supervised clustering algorithm.
  \item For instances of each cluster, we find the label whose mapped signature in the semantic visual space is the nearest one to the center of that cluster and assign that label to all of these instances.
\end{enumerate}

We use a simple ridge regression to map class signatures to visual features. We intend to find a mapping from class signatures to the deep visual representation space such that each mapped (seen) class signature is close to the samples of that class in this space in average.
The linear mapping is found using the following optimization problem:
\begin{equation} \label{eq:mapping}
  D = \argmin_D \norm{X_s - D Y_s} + \gamma \norm{D},
\end{equation}
where columns of $ Y_s \in \mathbb{R}^{r \times n_s} $ are the class signatures of the samples lied in the columns of $X_s$.
This optimization problem is known to have the following closed form solution:
\begin{equation} \label{eq:dic}
  D = X_s Y_s^T (Y_s Y_s^T + \gamma I)^{-1}.
\end{equation}
The parameter $\gamma$ is determined through cross validation as we will describe precisely in Section~\ref{experiments}.

Here, we intend to find labels for instances belonging to unseen classes. To this end, we want to find a clustering of instances in the space of deep visual features and assign a label to each cluster according to the distance between the center of that cluster and the mapped signature of the unseen classes (i.e., consider the label whose mapped signature is the closest one to the cluster center as the assigned label to the instances of this cluster). To find a better clustering of instances belonging to unseen classes, we can also incorporate labeled instances of seen classes too. The clustering problem over unseen instances, we encountered here, is different from the conventional semi-supervised learning problem \cite{chapel06}.
In fact, all labeled data are from seen classes and there is no labeled sample for unseen classes that is due to the special characteristic of zero-shot learning problem. Therefore, here, we propose a semi-supervised learning method which
can be seen as an extension to k-means suitable for this problem.
 We try to find a clustering such that labeled instances tend to be assigned to the corresponding classes and all instances tend to be close to the center of the clusters to which they are assigned:
\begin{equation} \label{eq:simple}
\minimize_{R, \mathbf{\mu_1, \ldots, \mu_k }}  \sum_{n,k} r_{nk} \lVert \mathbf{x_n - \mu_k} \rVert +
 \beta \sum_{n=1}^{N_s} \mathds{1}(\mathbf{r_n \neq z_n}),
\end{equation}
where $\mathbf{\mu_i'}$s are cluster centers and $R = [\mathbf{r_1, \ldots, r_{N_s + N_u }} ]$ is cluster assignments in one-hot encoding format.
The objective function is similar to that of the k-means clustering algorithm but for each labeled instance there is a penalty of $\beta$ if its assigned cluster number that is different from its label. Thus, this objective function encourages
the first $n_s$ clusters be corresponding to the seen classes.

Parameters $\beta$ and $k$ can be determined via the cross validation. However, in our experiments, we found out
the model is not very sensitive to these so we fix $\beta=1$
when data is normalized such that $\lVert x_i \rVert_1 = 1$. We set $k =  (n_s + n_u)$, i.e., the number of clusters is considered
 equal to the number of categories as a natural choice.

Finally, to assign labels to test instances, we use the mapping $D$ from Eq.\eqref{eq:dic} to
map class signatures to visual features, creating a set of \textit{class representatives}
 in the visual feature space. We then assign to all instances of a cluster the class label whose representative is
  the nearest to center of that cluster.



A key distinction between the clustering-based method presented here and other existing methods lies in the nature of the compatibility function. The compatibility function in
other works is a similarity measure between each instance and class description that is found independently for different instances.
 Here, the compatibility function relies strongly on the distribution of instances in the semantic space and the compatibility of a label for an instance is found according to the similarity of the cluster center to which this instance is assigned and the mapped signature of that label. Therefore, by considering the distribution of data points (via clustering)
  in designing the compatibility function we can reach a more reliable measure.
  This compatibility function can be plugged in every other method in this way that after final predictions are made by the method,
a clustering algorithm is ran on data and then we assign an identical label to all cluster
members by majority voting on those predictions. We found through experiment that this extra step will improve performance of
many existing methods.

Although the above method outperforms the state-of-the-art methods on most zero-shot recognition benchmarks, it uses only the instances of the seen classes to find the linear transformation from class signatures to the visual feature space and thus the proposed method may suffer from the domain shift problem introduced in \cite{eccv14}.
To overcome the domain shift problem more substantially, we propose an optimization problem for finding the linear transform from class signatures to the visual feature space that uses instances of both seen and unseen classes.

\subsection{Learning Mapping and Clustering Jointly}
\label{joint}
 In this section, we propose an optimization problem for learning a linear transformation from class signatures to the visual features space such that the mapped signatures are good representatives of the corresponding instances.
 We intend to learn a transformation such that for the seen classes, the sum of the squared distances of instances from the mapped signature of the corresponding class is minimized. Moreover, for instances of unseen classes, we can find class assignments such that the sum of the squared distances of unseen instances from the mapped signature of classes to which they are assigned is also minimized.
 The objective function is formulated as follows:
 \begin{align} \label{eq:main}
   \minimize_{R,D} \norm{X_s - D Y_s}  &+ \lambda \norm{X_u - D S_u R^T } + \gamma \norm{D} \\
   \text{s. t.} \quad & R \in \{0,1\}^{N_u \times n_u}. \nonumber
 \end{align}
The first term in the above optimization problem is identical to Eq.\eqref{eq:mapping} and the second one incorporates unlabeled data for learning the mapping $D$. By enforcing
 the signatures to be mapped close to test instances, this term confronts the domain shift problem. In fact, we seek a class assignment for instances of unseen classes such that we can learn a linear transformation on class signature to use the mapped signature of both seen and unseen classes as good representatives for the corresponding instances.
 The second term can be essentially considered as a clustering objective with two advantages. First, the number of clusters is no longer a
 parameter and it is determined by the number of unseen classes. Second, the cluster centers are set to be the mapped signatures of test classes.

\subsection{Optimization} \label{optimization}
%
Optimization of  the objective function in Eq.~\eqref{eq:simple} is done by alternating between
$\mathbf{\mu_i'}$s and R. $\mathbf{\mu_i'}$s are updated using:
\begin{equation}
  \mu_i = \frac{\sum_{n=1}^{N_s + N_u}  \mathds{1}(r_{ni}=1)\mathbf{x_n}}{\sum_{n=1}^{N_s+N_u}\mathds{1}(r_{ni}=1)},
\end{equation}
$R$ is updated by assigning each instance  to the cluster that minimizes the corresponding term in Eq.\eqref{eq:simple}.
To initialize $\mathbf{\mu_i'}$s, for clusters corresponding to seen classes the centers are set as mean of instances from that class. Centers of other
clusters are initialized using k-means++ \cite{kmeanspp} on unlabeled instances.
The overall training algorihm for LECA is presented in Algorithm \ref{leca}

The Eq.~\eqref{eq:main} is not convex and considering that $R$ is a partitioning of instances, the global optimization requires an
exhaustive search over all possible labeling of test data with $n_u$ labels. Therefore, we use a simple coordinate descent
method (like k-means). We alternate between optimizing $R$ and $D$ while fixing the other.
Having fixed the labeling $R$, the problem becomes a simple multi-task ridge regression which has the following closed-form solution:
\begin{equation} \label{eq:d_update}
  D = (X_s Y_s^T + \beta X_u R S_u^T) (Y_s Y_s^T + \beta S_u R^T R S_u^T  + \gamma I)^{-1}.
\end{equation}
By fixing $D$, the optimal $R$ can be achieved via assigning each instance to the closest class representative:
\begin{equation} \label{eq:r_update}
  r_{ij} = \mathds{1}[j = \argmin_{k} \lVert X_{u(i)} - D S_{u(k)} \rVert_2 ].
\end{equation}
Whenever a row of $R$ contains no 1's, i.e.  an empty cluster is encountered we assign 2\% of instances randomly to that cluster.
We continue alternating between updates of $D$ and $R$ till R remains constants, i.e., no label changes. In our experiments, this always happens
in less that 20 iterations.

To evade poor local minima, we propose a good initialization that is based on the simple method proposed in \ref{clustering}. We initialize  $R$ by final predictions found by this method.
\section{Experiments} \label{experiments}

\begin{table*}[ht]
\begin{minipage}{\textwidth}
\centering
\caption{Accuracy score (\%) of cluster assignments converted to labels
using majority voting on ground truth labels on four zero-shot recognition benchmarks.
Results are our method are average $\pm$ std of three runs.
} \vspace{2mm} \label{tab:cluster}
\begin{tabular}{|l|c|c|c|c|}
\hline
Clustering Method & Animals with Attributes & CUB-2011 & aPascal-aYahoo & SUN Attribute \\
\hline
k-means                             &  65.80                 & 35.61           & 65.37                & 17.49    \\
\hline
Ours (Simple)                     & \textbf{70.74$\pm$0.32}  & \textbf{42.63$\pm$0.07} & \textbf{69.93$\pm$ 3.4} & \textbf{ 45.50$\pm$1.32} \\
\hline
\end{tabular}
\vspace{2mm}
\end{minipage}
\end{table*}

\begin{table*}[ht]
\begin{minipage}{\textwidth}
\centering
\caption{Classification accuracy in \% on four public datasets: Animals with Attributes, CUB-2011, aPascal-aYahoo and SUN
in form of average $\pm$ std.
} \vspace{2mm} \label{tab:results}
\begin{tabular}{|l|l|c|c|c|c|}
\hline
Feature & Method & Animals with Attributes & CUB-2011 & aPascal-aYahoo & SUN \\
\hline
{Shallow}
& Li and Guo \cite{li15max}                 &  38.2$\pm$2.3   &                 &                         & 18.9$\pm$2.5 \\
& Li \etal~\cite{semi15}                    &  40.05$\pm$2.25 &                 &   24.71 $\pm$3.19       &     \\
& Jayaraman and Grauman \cite{jayaraman14}  &43.01 $\pm$ 0.07 &                 & 26.02 $\pm$ 0.05        & 56.18 $\pm$ 0.27 \\
\hline
{GoogleNet}
& Akata \etal~\cite{Akata2015}              & 66.7            & 50.1            &                         & \\
& Changpinyo \etal~\cite{Synthesized}       & 72.9            & 54.5            &                         & 62.7 \\
& Xian \etal~\cite{Xian2016}                & 71.9            & 45.5            &                         & \\
\hline
{VGG-19}
& Khodirov \etal \cite{Kodirov2015}
                                            & 73.2            &  39.5           & 26.5                    &  \\
& Akata \etal~\cite{Akata2015}              & 61.9            &  50.1           &                         & \\
& Zhang and Saligrama \cite{sse}            &  76.33$\pm$0.53 & 30.41 $\pm$0.20 &   46.23 $\pm$ 0.53      & 82.50 $\pm$ 1.32    \\
& Zhang and Saligrama \cite{agnostic}       &  80.46$\pm$0.53 & 42.11 $\pm$0.55 &   \textbf{50.35 $\pm$ 2.97}      & 83.83 $\pm$ 0.29    \\

& Ours (Simple)                             & 86.58$\pm$1.12               & 52.19$\pm$0.83              & 49.86$\pm$2.36              & 84.50$\pm$1.32 \\
& Ours (Joint - init D)                     & 83.03                        & 57.55                       & 42.62          & 72.50\\
& Ours (Joint - init R)                     & \textbf{\em 88.64$\pm$0.04}  & \textbf{\em 58.80$\pm$0.64} & 49.77$\pm$2.02 & \textbf{\em 86.16$\pm$0.57} \\
\hline
\end{tabular}
\end{minipage}\vspace{-3mm}
\end{table*}
In this section, we conduct experiments on the popular benchmarks to obtain results of the proposed method on these benchmarks and compare them with those of the other methods.

\textbf{Datasets.}
We evaluate our proposed methods on four popular public benchmarks for zero-shot classification.
(1) Animal with Attributes (AwA) \cite{lampert09}. There are images of 50 mammal species in this data set
Each class is described by a single $85-$dimensional attribute vector. We use the continuous attributes rather than
the binary version as it has proved to be more discriminative in previous works like \cite{Akata2015}. The train/test split provided by the dataset is used accordingly.
(2) aPascal/aYahoo \cite{farhadi09}. The 20 categories from Pascal VOC 2008 \cite{pascal} are considered as seen classes and
categories from aYahoo are considered to be unseen. As this dataset provides instance level attribute vectors,
for class signatures we use the average of the provided instance attributes.
(3) SUN Attribute \cite{sun}. The dataset consists of 717 categories and all images are annotated with 102 attributes, we just
use the average attributes among all instances of each categories for our experiments. We use the same train/test spilt
as in \cite{jayaraman14} where 10 classes have been considered unseen.
(4) Caltech UCSD Birds-2011 (CUB) \cite{cub}. This a dataset for fine-grained classification task. There are 200 species of
birds where each image has been annotated with 312 binary attributes. Again, we average over instances to get continuous class signatures.
We use the same train/test split as in \cite{akata13} (and many other following works) to make comparison possible.

As our method relies on meaningful structure in visual features domain, we use features from a deep CNN known that are
 more discriminative than \textit{shallow} features like SIFT or HOG. We report results using
  $4096-$dimensional features from the first fully connected layer of 19 layer VGG network \cite{vgg}
pre-trained on the subset, provided publicly by \cite{sse}.

\textbf{Testing Cluster Assumption:}
First, to give evidence for our key assumption of our method that instances from each class usually form a cluster in visual feature domains
and to demonstrate effectiveness of our proposed clustering algorithm we design an
experiment in which instances from unseen categories are clustered using our proposed clustering algorithm and also the k-means algorithm. Then,
each cluster is assigned with a class label based on majority voting on ground truth labels. The number of clusters
 is set to the number of classes as a natural choice (increasing the number of clusters improves the accuracy).
For the k-means algorithm, we use the implementation available in Scikit-learn library \cite{scikit-learn} and run it with 20 different initializations
and report results of that one with the best score.
Accuracy of this labeling scheme that is based on clustering is reported in Table~\ref{tab:cluster}.
These results shows the effectiveness of our proposed clustering method and that the cluster structure assumption in the visual semantic space is usually right.

\textbf{Cross Validation:}
To adjust parameters $\gamma$ and $\beta$ in Eq.~\ref{eq:d_update}, we split training data into train and validation sets.
We choose a number of categories randomly from training data as validation categories. For each data set, the size of the
validation set has the same ratio to the train set as the size of the test categories to the total of the train and the validation one.
In our experiments, we used $10-$fold cross validation, i.e., average results from ten different validation splits are used to decide on
optimal parameters.
Once optimal $\gamma$ and $\beta$ are determined through the grid search by testing on validation set, the model
is then trained on all seen categories.

We summarize our experimental results in Table~\ref{tab:results}.
\textit{Ours (Simple)} corresponds to the method presented in Section~\ref{clustering}.
\textit{Ours(init D)} and \textit{Ours(init - R)} correspond to optimizing Eq.~\eqref{eq:main}
with respectively initializing $D$ using Eq.~\eqref{eq:dic} and initializing $R$ by our simple method proposed in Section~\ref{clustering}.
For our methods, average and standard deviation of different runs are reported. As it can be seen, the initialization done
by our simple method has critical effect on the performance. This can be justified by noting the information from structure of
unlabeled data is leveraged when initializing $R$ while such information is absent in initializing $D$.

 For other methods, we use the results reported in their original publication. Note that some experimental settings of these works may differ from those of ours. We did not re-implement any of the other methods and if the original paper does not report results on a data set we leave the corresponding cell as blank.
Our method performs the best on three out of the four datasets (outperforms the others on all except to the aPascal-aYahoo dataset). This can be explained by the nature of the dataset in which class signatures obtained by averaging instance attributes are very similar. We suppose trying to learn
more discriminative signatures from data can potentially improve the result. We investigate this in our future work.


\section{Conclusion} \label{conclusion}
In this paper, we proposed semi-supervised methods for zero-shot object recognition. We used the space of deep visual features as a semantic visual space and learned a linear transformation to map class signatures to this space such that the mapped signatures provide good representative of the corresponding instances. We utilized this property that the rich deep visual features provide a representation space in which samples of each class are usually condensed in a cluster. In the proposed method that jointly learns the mapping of class signatures and the class assignments of unlabeled data, we used also unlabeled instances of unseen classes when learning the mapping to alleviate the domain shift problem. Experimental results showed that the proposed method generally outperformed the other recent methods.
{\small
\bibliographystyle{ieee}
\bibliography{semi_suepervised_zsl_by_clustering.bib}

\begin{thebibliography}{10}\itemsep=-1pt

\bibitem{Akata2016}
Z.~Akata, M.~Malinowski, M.~Fritz, and B.~Schiele.
\newblock {Multi-Cue Zero-Shot Learning with Strong Supervision}.
\newblock {\em arXiv preprint arXiv:1603.08754}, 2016.

\bibitem{akata13}
Z.~Akata, F.~Perronnin, Z.~Harchaoui, and C.~Schmid.
\newblock Label-embedding for attribute-based classification.
\newblock In {\em Computer Vision and Pattern Recognition (CVPR), IEEE
  Conference on}, pages 819--826, 2013.

\bibitem{Akata2015pami}
Z.~Akata, F.~Perronnin, Z.~Harchaoui, and C.~Schmid.
\newblock Label-embedding for image classification.
\newblock {\em IEEE Transactions on Pattern Analysis and Machine Intelligence},
  PP(99), 2015.

\bibitem{Akata2015}
Z.~Akata, S.~Reed, D.~Walter, H.~Lee, and B.~Schiele.
\newblock {Evaluation of Output Embeddings for Fine-Grained Image
  Classification}.
\newblock In {\em Computer Vision and Pattern Recognition (CVPR), IEEE
  Conference on}, 2015.

\bibitem{kmeanspp}
D.~Arthur and S.~Vassilvitskii.
\newblock k-means++: the advantages of careful seeding.
\newblock In {\em In Proceedings of the eighteenth annual ACM-SIAM symposium on
  Discrete algorithms}, pages 1027--1035, 2007.

\bibitem{ba2015}
J.~Ba, K.~Swersky, S.~Fidler, and R.~Salakhutdinov.
\newblock {Predicting Deep Zero-Shot Convolutional Neural Networks using
  Textual Descriptions}.
\newblock {\em arXiv preprint arXiv:1506.00511}, 2015.

\bibitem{Synthesized}
S.~Changpinyo, W.~Chao, B.~Gong, and F.~Sha.
\newblock Synthesized classifiers for zero-shot learning.
\newblock {\em CoRR}, abs/1603.00550, 2016.

\bibitem{chapel06}
O.~Chapelle, B.~Sch{\"o}lkopf, and A.~Zien.
\newblock {\em Semi-Supervised Learning}.
\newblock MIT Press, Cambridge, MA, 2006.

\bibitem{mohamed13}
M.~Elhoseiny, B.~Saleh, and A.~Elgammal.
\newblock Write a classifier: Zero-shot learning using purely textual
  descriptions.
\newblock In {\em Computer Vision (ICCV), IEEE Conference on}, pages
  2584--2591, 2013.

\bibitem{farhadi09}
A.~Farhadi, I.~Endres, D.~Hoiem, and D.~Forsyth.
\newblock {Describing Objects by Their Attributes}.
\newblock In {\em Computer Vision and Pattern Recognition (CVPR), IEEE
  Conference on}, pages 1778--1785, 2009.

\bibitem{devise}
A.~Frome, G.~S. Corrado, J.~Shlens, S.~Bengio, J.~Dean, M.~Ranzato, and
  T.~Mikolov.
\newblock {DeViSE: A Deep Visual-Semantic Embedding Model}.
\newblock In {\em Advances in Neural Information Processing Systems (NIPS) 26},
  pages 2121--2129, 2013.

\bibitem{eccv14}
Y.~Fu, T.~M. Hospedales, T.~Xiang, Z.~Fu, and S.~Gong.
\newblock Transductive multi-view embedding for zero-shot recognition and
  annotation.
\newblock In {\em Computer Vision (ECCV), European Conference on}, volume 6315.
  2014.

\bibitem{Fu2016}
Y.~Fu and L.~Sigal.
\newblock {Semi-supervised Vocabulary-informed Learning}.
\newblock {\em arXiv preprint arXiv:1604.07093}, 2016.

\bibitem{pascal}
D.~Hoiem, S.~K. Divvala, and J.~H. Hays.
\newblock Pascal voc 2008 challenge, 2008.

\bibitem{jayaraman14}
D.~Jayaraman and K.~Grauman.
\newblock {Zero-shot recognition with unreliable attributes}.
\newblock In {\em Advances in Neural Information Processing Systems (NIPS) 27},
  pages 3464--3472. 2014.

\bibitem{Kodirov2015}
E.~Kodirov, T.~Xiang, Z.~Fu, and S.~Gong.
\newblock {Unsupervised Domain Adaptation for Zero-Shot Learning}.
\newblock In {\em Computer Vision (ICCV), IEEE Conference on}, pages
  2927--2936, 2015.

\bibitem{lampert09}
C.~Lampert, H.~Nickisch, and S.~Harmeling.
\newblock Learning to detect unseen object classes by between-class attribute
  transfer.
\newblock In {\em Computer Vision and Pattern Recognition (CVPR), IEEE
  Conference on}, pages 951--958, 2009.

\bibitem{bengio08}
H.~Larochelle, D.~Erhan, and Y.~Bengio.
\newblock {Zero-data learning of new tasks}.
\newblock In {\em National Conference on Artificial Intelligence (AAAI)}, pages
  646--651, 2008.

\bibitem{li15max}
X.~Li and Y.~Guo.
\newblock Max-margin zero-shot learning for multi-class classification.
\newblock In {\em Proceedings of the Eighteenth International Conference on
  Artificial Intelligence and Statistics (AISTATS)}, pages 626--634, 2015.

\bibitem{ajoint11}
D.~Mahajan, S.~Sellamanickam, and V.~Nair.
\newblock A joint learning framework for attribute models and object
  descriptions.
\newblock In {\em Computer Vision (ICCV), IEEE International Conference on},
  pages 1227--1234, 2011.

\bibitem{word2vec}
T.~Mikolov, I.~Sutskever, K.~Chen, G.~S. Corrado, and J.~Dean.
\newblock Distributed representations of words and phrases and their
  compositionality.
\newblock In {\em Advances in Neural Information Processing Systems (NIPS) 26},
  pages 3111--3119. 2013.

\bibitem{convex}
M.~Norouzi, T.~Mikolov, S.~Bengio, Y.~Singer, J.~Shlens, A.~Frome, G.~Corrado,
  and J.~Dean.
\newblock Zero-shot learning by convex combination of semantic embeddings.
\newblock In {\em International Conference on Learning Representations}, 2014.

\bibitem{hinton09}
M.~Palatucci, G.~Hinton, D.~Pomerleau, and T.~M. Mitchell.
\newblock Zero-shot learning with semantic output codes.
\newblock In {\em Advances in Neural Information Processing Systems (NIPS) 22},
  pages 1410--1418. 2009.

\bibitem{sun}
G.~Patterson, C.~Xu, H.~Su, and J.~Hays.
\newblock The sun attribute database: Beyond categories for deeper scene
  understanding.
\newblock {\em International Journal of Computer Vision}, 108(1-2):59--81,
  2014.

\bibitem{scikit-learn}
F.~Pedregosa, G.~Varoquaux, A.~Gramfort, V.~Michel, B.~Thirion, O.~Grisel,
  M.~Blondel, P.~Prettenhofer, R.~Weiss, V.~Dubourg, J.~Vanderplas, A.~Passos,
  D.~Cournapeau, M.~Brucher, M.~Perrot, and E.~Duchesnay.
\newblock Scikit-learn: Machine learning in {P}ython.
\newblock {\em Journal of Machine Learning Research}, 12:2825--2830, 2011.

\bibitem{pennington2014glove}
J.~Pennington, R.~Socher, and C.~D. Manning.
\newblock Glove: Global vectors for word representation.
\newblock In {\em Empirical Methods in Natural Language Processing (EMNLP)},
  pages 1532--1543, 2014.

\bibitem{Qiao2016}
R.~Qiao, L.~Liu, C.~Shen, and A.~van~den Hengel.
\newblock {Less is more: zero-shot learning from online textual documents with
  noise suppression}.
\newblock {\em arXiv preprint arXiv:1604.01146}, 2016.

\bibitem{emb15}
B.~Romera-Paredes and P.~H.~S. Torr.
\newblock { An Embarrassingly Simple Approach to Zero-shot Learning}.
\newblock {\em Journal of Machine Learning Research}, 37, 2015.

\bibitem{semi15}
D.~Schuurmans and A.~B. Tg.
\newblock {Semi-Supervised Zero-Shot Classification with Label Representation
  Learning}.
\newblock In {\em Computer Vision (ICCV), IEEE Conference on}, 2015.

\bibitem{Akata2016rnn}
B.~S. {Scott Reed, Zeynep Akata, Honglak Lee}.
\newblock {Learning Deep Representations of Fine-Grained Visual Descriptions}.
\newblock {\em CVPR}, 2016.

\bibitem{vgg}
K.~Simonyan and A.~Zisserman.
\newblock Very deep convolutional networks for large-scale image recognition.
\newblock {\em CoRR}, 2014.

\bibitem{suzuki14}
M.~Suzuki, H.~Sato, S.~Oyama, and M.~Kurihara.
\newblock Transfer learning based on the observation probability of each
  attribute.
\newblock In {\em Systems, Man and Cybernetics (SMC), IEEE International
  Conference on}, pages 3627--3631, 2014.

\bibitem{cub}
C.~Wah, S.~Branson, P.~Welinder, P.~Perona, and S.~Belongie.
\newblock {The Caltech-UCSD Birds-200-2011 Dataset}.
\newblock Technical report, 2011.

\bibitem{unified13}
X.~Wang and Q.~Ji.
\newblock A unified probabilistic approach modeling relationships between
  attributes and objects.
\newblock In {\em Computer Vision (ICCV), IEEE International Conference on},
  pages 2120--2127, 2013.

\bibitem{Xian2016}
Y.~Xian, Z.~Akata, G.~Sharma, Q.~Nguyen, M.~Hein, and B.~Schiele.
\newblock {Latent Embeddings for Zero-shot Classification}.
\newblock {\em arXiv preprint arXiv:1603.08895}, mar 2016.

\bibitem{Yu2013}
F.~X. Yu, L.~Cao, R.~S. Feris, J.~R. Smith, and S.-F. Chang.
\newblock {Designing Category-Level Attributes for Discriminative Visual
  Recognition}.
\newblock In {\em Computer Vision and Pattern Recognition (CVPR), IEEE
  Conference on}, pages 771--778, 2013.

\bibitem{topicmodel}
X.~Yu and Y.~Aloimonos.
\newblock Attribute-based transfer learning for object categorization with
  zero/one training example.
\newblock In {\em Computer Vision (ECCV), European Conference on}, volume 6315,
  pages 127--140. 2010.

\bibitem{agnostic}
Z.~Zhang and V.~Saligrama.
\newblock Zero-shot learning via joint latent similarity embedding.
\newblock {\em arXiv preprint arXiv:1511.04512}, 2015.

\bibitem{sse}
Z.~Zhang and V.~Saligrama.
\newblock {Zero-Shot Learning via Semantic Similarity Embedding}.
\newblock In {\em Computer Vision (ICCV), IEEE Conference on}, 2015.

\end{thebibliography}
}

\end{document}